\documentclass{article}

\usepackage{PRIMEarxiv}

\usepackage[utf8]{inputenc} 
\usepackage[T1]{fontenc}    
\usepackage{hyperref}       
\usepackage{url}            
\usepackage{booktabs}       
\usepackage{amsmath,amssymb} 
\usepackage{amsfonts}       
\usepackage{nicefrac}       
\usepackage{microtype}      
\usepackage{fancyhdr}       
\usepackage{graphicx}       
\graphicspath{{media/}}     

\title{UniCompress: From Multi-LoRA Experts to One Compact Model for Edge Intelligence
}

\author{
  Mingkang Liu \\
  Juhaokan Technology Co., Ltd. \\
  Qingdao, China \\
  \texttt{liumingkang@hisense.com} \\
  \And
  Peipei Jiao \\
  Juhaokan Technology Co., Ltd. \\
  Qingdao, China \\
  \And
  Weiming Meng \\
  Juhaokan Technology Co., Ltd. \\
  Qingdao, China \\
  \And
  Tianyuan Liu \\
  Juhaokan Technology Co., Ltd. \\
  Qingdao, China \\
  \And
  Lichen Pang\thanks{Corresponding author} \\
  Juhaokan Technology Co., Ltd. \\
  Qingdao, China \\
  \texttt{cliff.pan@hisense.com} \\
  \And
  Junyan Li \\
  Juhaokan Technology Co., Ltd. \\
  Qingdao, China \\
}

\begin{document}
\maketitle

\begin{abstract}
Embodied and edge systems increasingly rely on vision--language models (VLMs) to support heterogeneous skills, yet deploying a separate task adapter for each skill while retaining a high-precision backbone creates substantial memory and management costs. We present \textbf{UniCompress}, a deployment-oriented framework that converts multi-task expertise into one compact, adapter-free model. Starting from a consolidated BF16 Qwen3-VL-4B teacher, UniCompress constructs an asymmetric INT2 student using GPTQ initialization and teacher-guided quantization recovery. The approach jointly quantizes and recovers the projection weights in the vision encoder and language backbone, enabling aggressive compression of the complete vision--language model while preserving its multi-task behavior. We evaluate UniCompress on four heterogeneous embodied tasks: posture recognition, household-safety understanding, multimodal dialogue routing, and mobile-GUI action prediction. UniCompress reduces the approximate weight storage from 8.9 GB in BF16 to 2.1 GB in INT2, corresponding to a reduction of approximately 76\%. It improves the INT2 macro task score by 34.3 percentage points over GPTQ-INT2 while remaining within 0.9 points of the consolidated BF16 teacher. These results demonstrate the potential of teacher-guided joint vision--language INT2 recovery for deploying a single multi-skill VLM on resource-constrained embodied platforms.
\end{abstract}

\section{Introduction}

Large foundation models, including large language models (LLMs) and vision--language models (VLMs), have demonstrated strong capabilities in perception, reasoning, interaction, and decision-making. These advances are enabling a new generation of embodied and edge-intelligence applications in which an intelligent agent is expected to perform multiple heterogeneous skills within a shared system. However, deploying such models on resource-constrained devices remains challenging because of their substantial memory footprint, computational cost, and energy consumption. Practical embodied intelligence therefore requires a compact model that can preserve diverse task capabilities while satisfying strict deployment constraints.

Parameter-efficient fine-tuning adapts foundation models without updating or storing a separate backbone for every downstream task. Low-Rank Adaptation (LoRA)~\cite{lora} introduces lightweight trainable modules while keeping the pretrained backbone frozen, and Multi-LoRA systems extend this design to multiple skills by attaching task-specific adapters to a shared model. However, the deployment overhead grows with the number of tasks: every adapter must be stored, selected, and managed at runtime, which complicates model versioning, task switching, and deployment on platforms designed around a single static checkpoint. These limitations motivate consolidating multiple task-specific experts into one unified model.

Adapter consolidation alone, however, does not reduce the memory cost of the shared backbone, which dominates the deployment footprint. Low-bit quantization methods such as GPTQ~\cite{gptq} and AWQ~\cite{awq} address this cost: at moderate bit widths such as INT4, post-training quantization preserves most of the original performance, whereas extreme INT2 quantization introduces severe approximation errors. For a multi-task model, these errors must be controlled across all capabilities simultaneously, and gains on dominant tasks may come at the expense of less frequent or more sensitive skills. The problem is further amplified for VLMs when both the visual and language components are quantized. Preserving consolidated multi-task knowledge under extreme joint vision--language compression therefore remains an important yet underexplored problem.

To address these challenges, we propose \textbf{UniCompress}, a unified framework that transforms distributed task expertise into one compact model for efficient edge intelligence. UniCompress follows a deployment-oriented compression process. It first represents the knowledge of multiple task-specific adaptations within a single consolidated multi-task teacher, eliminating the need to maintain separate adapters during deployment. It then transfers this consolidated knowledge to an extremely low-bit student through teacher-guided quantization recovery, jointly quantizing the projection weights in the vision encoder and language backbone. Rather than treating quantization solely as a local weight-reconstruction problem, UniCompress preserves task behavior by aligning the student with the output distributions of the high-precision teacher during training. The resulting model is exported as a single INT2 checkpoint and requires neither multiple task adapters nor runtime adapter selection. Although INT4 quantization substantially reduces model size, its memory footprint may still exceed the budgets of highly resource-constrained embodied and edge devices. We therefore target INT2 deployment, which offers a more aggressive compression ratio but poses substantially greater challenges in preserving multi-task capabilities.

The main contributions of this work are summarized as follows: (i) we formulate the deployment of multi-task foundation models as a unified compression problem that jointly considers adapter management overhead and the dominant memory cost of the shared vision--language backbone; (ii) we propose \textbf{UniCompress}, a deployment-oriented framework that consolidates diverse task expertise into one model and applies teacher-guided joint vision--language INT2 recovery to preserve multi-task capabilities under extreme quantization; and (iii) experiments on four on-device tasks with Qwen3-VL-4B show that UniCompress reduces the approximate weight storage from 8.9 GB to 2.1 GB and recovers 34.3 percentage points over GPTQ-INT2, within 0.9 points of the consolidated BF16 model.

\begin{figure*}[t]
\centerline{\includegraphics[width=0.85\textwidth]{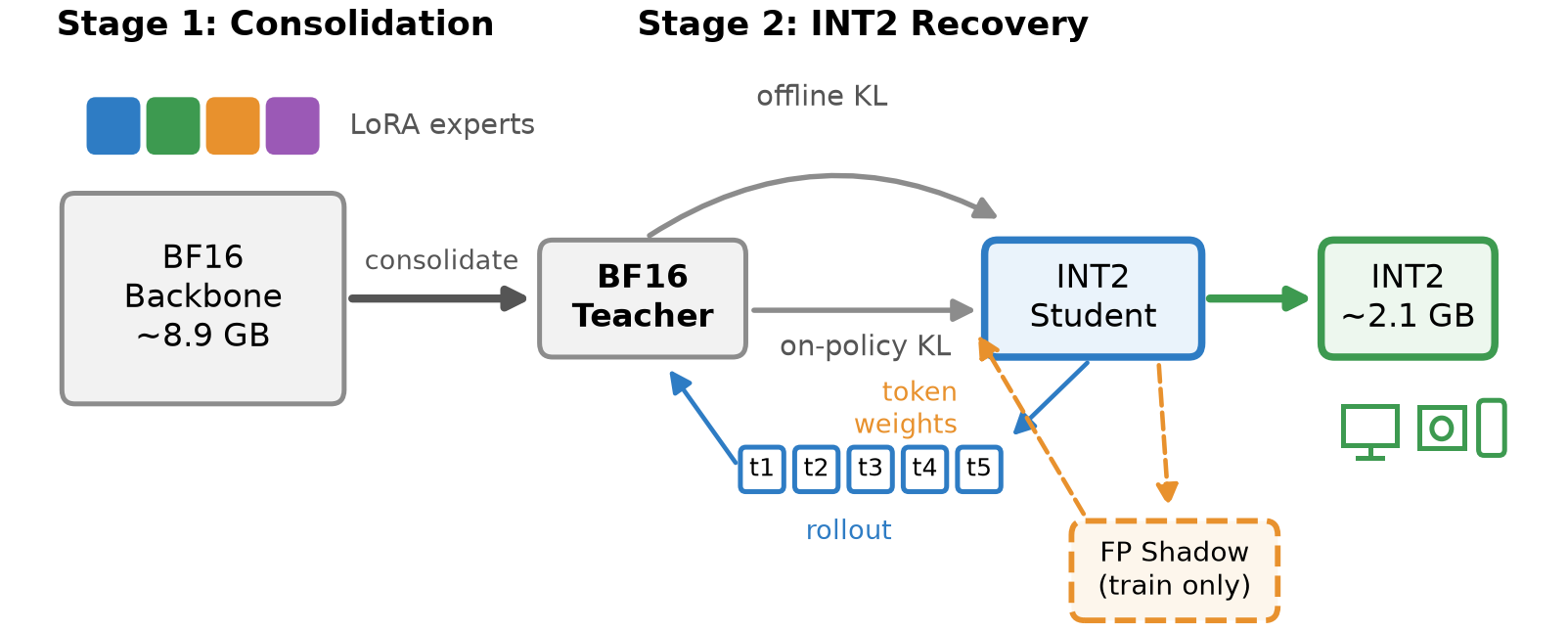}}
\caption{Overview of UniCompress. Stage~1 consolidates four task-specific LoRA experts into a single Unified-BF16 teacher. Stage~2 trains a deployment-faithful INT2 student with offline anchoring and on-policy recovery, where the floating-point shadow policy assigns token weights according to quantization sensitivity. Only the INT2 student is deployed.}
\label{fig:framework}
\end{figure*}

\section{Related Work}
\label{sec:related_work}

\paragraph{Multi-Task Adaptation}
Parameter-efficient fine-tuning methods~\cite{houlsby2019adapter,li2021prefixtuning} adapt foundation models without updating the complete backbone. LoRA~\cite{lora} represents task-specific updates with low-rank matrices and has become a common choice for constructing multi-skill systems. A Multi-LoRA deployment~\cite{sheng2024slora,chen2024punica} can share one backbone across tasks, but still requires multiple adapters to be stored, selected, and managed at runtime. Adapter merging~\cite{ilharco2023taskarithmetic,yadav2023ties} removes this runtime switching cost by consolidating task-specific updates into a single checkpoint. In this work, we start from such a consolidated multi-task teacher and focus on the complementary problem of compressing it into one adapter-free model suitable for edge deployment.

\paragraph{Low-Bit Quantization of Vision-Language Models}
Post-training quantization methods such as GPTQ~\cite{gptq} and AWQ~\cite{awq} reduce the memory footprint of large models by optimizing weight reconstruction or preserving salient weights. Quantization-aware approaches such as LLM-QAT~\cite{liu2023llmqat} further use training-time quantization and distillation to recover performance at reduced precision. For VLMs, Q-VLM~\cite{wang2024qvlm} models cross-layer dependencies during quantization, while MBQ~\cite{li2024mbq} accounts for the different sensitivities of visual and textual tokens. Existing studies mainly focus on post-training calibration or moderate W3/W4 settings evaluated on general-purpose benchmarks. Efforts toward more aggressive regimes, such as 1-bit weight parameterization~\cite{ma2024bitnet} demonstrate the feasibility of extreme compression but remain limited to language-only settings. In contrast, we target asymmetric INT2 compression of a consolidated multi-task VLM and evaluate it on structured edge tasks whose outputs include safety labels, multimodal routing decisions, and GUI actions. At this precision, quantization can alter early predictions and consequently change the subsequent autoregressive states, making local weight reconstruction insufficient for behavior recovery.

\paragraph{On-Policy Distillation}
Conventional distillation~\cite{hinton2015distilling,kim2016sequence} typically supervises the student on ground-truth or teacher-generated prefixes, which may differ from the states encountered during student inference. MiniLLM~\cite{gu2024minillm} optimizes reverse KL divergence using student samples, DistiLLM~\cite{ko2024distillm} improves its training efficiency with skewed divergence objectives, while GKD~\cite{agarwal2024onpolicy} queries the teacher on student-generated trajectories to reduce this train--inference mismatch. Recent work such as vOPD~\cite{oh2026vopd} further improves the stability of sampled on-policy gradients through variance reduction. These methods primarily address capacity differences between normally parameterized teacher and student models. Our setting instead considers a student policy directly perturbed by deployment-level quantization. UniCompress combines offline knowledge anchoring with trajectories generated by the actual INT2 student and uses a parameter-sharing floating-point shadow policy to emphasize quantization-sensitive states.

\section{Method}

\label{sec:method}

\subsection{Overview}
\label{sec:overview}

We study the recovery of an ultra-low-bit vision-language model under realistic on-device inference. The two stages of UniCompress play strictly complementary roles: Stage~1 (LoRA consolidation) removes the \emph{management} cost of multiple adapters by merging them into a single BF16 model without changing precision or backbone memory---and, because one weight set then serves all tasks, additionally lets the KV cache of an ongoing session be inherited across tasks, which per-task adapter switching would invalidate---whereas Stage~2 (INT2 recovery) removes the \emph{backbone memory} cost by converting that consolidated model into a mixed-precision asymmetric INT2 student while preserving its multi-task behavior. Unlike conventional quantization methods that optimize the student only on fixed training sequences, our method explicitly considers the autoregressive state distribution visited by the deployed quantized model. A small perturbation of an early-token distribution may change the generated token and consequently move the model to prefixes that are absent from the offline training data. Such errors become particularly severe at 2-bit precision.

Fig.~\ref{fig:framework} illustrates the overall pipeline. Our framework contains a frozen BF16 teacher policy $p_T$, a deployment-form quantized student policy $q_\theta^Q$, and a parameter-sharing floating-point shadow policy $q_\theta^F$. The teacher and student use the same Qwen3-VL-4B architecture. The student first learns from annotated offline sequences to preserve task knowledge. It then generates its own trajectories using the actual INT2 fake-quantized projection weights in both the vision encoder and language backbone, on which the teacher provides token-level supervision. The shadow policy is used only during training to identify states whose predictions are strongly affected by quantization. The final deployed model contains only the INT2 student and introduces no additional inference module.

\subsection{Deployment-Faithful INT2 Student}
\label{sec:int2_student}

The teacher and student share the same vision-language architecture. The resulting student is a mixed-precision model: we quantize the vision-encoder projection weights with group-wise asymmetric INT4 and the language-backbone projection weights with group-wise asymmetric INT2, while keeping the embeddings and normalization layers in BF16. The trainable BF16 master weights are initialized from a calibrated GPTQ model and dynamically fake-quantized~\cite{jacob2018quantization} during every student forward pass.

For a weight group $W_g$ with bit width $b=2$, we use group-wise asymmetric quantization with group size 64. Its scale and zero point are

\begin{equation}
s_g =
\frac{\max(W_g)-\min(W_g)}{2^b-1},
\end{equation}

\begin{equation}
z_g =
\operatorname{clip}_{[0,\,2^b-1]}\!\left(
\operatorname{round}\!\left(
-\frac{\min(W_g)}{s_g}
\right)
\right).
\end{equation}

The corresponding fake-quantized weight is

\begin{equation}
\widehat W_g =
s_g\left[
\operatorname{clip}_{[0,\,2^b-1]}\!\left(
\operatorname{round}\!\left(W_g/s_g\right)+z_g
\right)-z_g
\right].
\label{eq:fake_quant}
\end{equation}

We apply a straight-through estimator~\cite{bengio2013ste} to the rounding operation and update only the master projection weights. Both the visual and language projection weights are fake-quantized during training forwards and student rollouts. Therefore, the generated trajectories reflect the complete quantized vision--language policy that will actually be deployed on the edge device.

\subsection{Offline Knowledge Anchoring}
\label{sec:offline}

Given a multimodal prompt $x$ and an annotated response
$y^\ast=(y_1^\ast,\ldots,y_N^\ast)$, we first distill the teacher on the ground-truth prefixes
$h_t^\ast=(x,y_{<t}^\ast)$. Let $p_t$ and $q_t$ denote the teacher and quantized-student token distributions at $h_t^\ast$. We use the teacher probability of the target token, $c_t = p_T(y_t^\ast \mid h_t^\ast)$, to balance forward and reverse KL divergences. The offline anchoring loss is

\begin{equation}
\mathcal{L}_{\mathrm{off}} =
\frac{1}{N}\sum_{t=1}^{N}
\left[
(1-c_t)D_{\mathrm{KL}}(p_t\Vert q_t)
+
c_tD_{\mathrm{KL}}(q_t\Vert p_t)
\right].
\label{eq:offline}
\end{equation}

This confidence-aware objective preserves the teacher's broad distribution when its target confidence is low, while encouraging mode-seeking behavior on confident tokens. We use it as a knowledge-preservation term rather than as the main contribution of our method.

\subsection{Quantized-Policy On-Policy Recovery}
\label{sec:opd}

Offline distillation does not expose the student to prefixes caused by its own quantization errors. We therefore generate a trajectory $\widetilde y \sim q_\theta^Q(\cdot\mid x)$ directly from the jointly quantized vision--language policy, supervising the student on these self-generated trajectories through on-policy distillation (OPD).

For every generated prefix
$\widetilde h_t=(x,\widetilde y_{<t})$, we evaluate both the frozen teacher and the quantized student. The basic on-policy objective is the token-level reverse KL divergence

\begin{equation}
\ell_t^{\mathrm{on}} =
D_{\mathrm{KL}}
\left(
q_\theta^Q(\cdot\mid\widetilde h_t)
\Vert
p_T(\cdot\mid\widetilde h_t)
\right).
\label{eq:online_token}
\end{equation}

The discrete rollout is treated as fixed when computing gradients. We use the analytic full-vocabulary divergence in Eq.~\eqref{eq:online_token}, avoiding the high gradient variance of single-token Monte Carlo estimators. Thus, the online branch directly corrects the behavior of the quantized student at the states it is likely to encounter during deployment.

\subsection{Quantization-Induced Shift Reweighting}
\label{sec:shift_weight}

Not every rollout state is equally affected by INT2 quantization. To separate quantization-induced errors from general teacher--student differences, we construct a floating-point shadow policy $q_\theta^F$ by temporarily bypassing fake quantization in both the visual and language projection weights while sharing exactly the same master weights with $q_\theta^Q$. The shadow policy introduces no additional trainable parameters.

At each quantized-student prefix, we measure the deployment discrepancy as

\begin{equation}
d_t =
D_{\mathrm{JS}}
\left(
q_\theta^Q(\cdot\mid\widetilde h_t),
q_\theta^F(\cdot\mid\widetilde h_t)
\right).
\label{eq:shift}
\end{equation}

A larger $d_t$ indicates that the state is more sensitive to joint vision--language INT2 quantization. We normalize the discrepancy within each trajectory:

\begin{equation}
w_t =
(1-\beta)
+
\beta T
\frac{\exp(d_t/\tau)}
{\sum_{j=1}^{T}\exp(d_j/\tau)},
\label{eq:weight}
\end{equation}

where $\tau$ controls the concentration and $\beta$ balances uniform and discrepancy-aware supervision. The weights are detached during optimization. The resulting on-policy loss is

\begin{equation}
\mathcal{L}_{\mathrm{on}} =
\frac{1}{T}
\sum_{t=1}^{T}
\operatorname{sg}(w_t)\,
\ell_t^{\mathrm{on}},
\label{eq:weighted_online}
\end{equation}

where $\operatorname{sg}(\cdot)$ denotes stop-gradient. This design concentrates recovery on prefixes whose behavior is altered most strongly by the deployment quantizer, rather than uniformly distilling all generated tokens.

Finally, the student is optimized with

\begin{equation}
\mathcal{L} =
\mathcal{L}_{\mathrm{off}}
+
\lambda(s)\mathcal{L}_{\mathrm{on}},
\label{eq:total}
\end{equation}

where $\lambda(s)$ is linearly increased during an initial warm-up period. This allows the offline branch to establish a stable task anchor before on-policy recovery becomes dominant. The teacher and floating-point shadow policy are discarded after training; consequently, the exported model retains standard joint vision--language INT2 inference with no additional memory or latency overhead.

\section{Experiments}
\label{sec:experiments}

We evaluate both stages of UniCompress. First, we examine whether multiple task-specific LoRA experts can be consolidated into a single BF16 model without losing their specialized capabilities. Second, we evaluate whether the consolidated model can retain these capabilities after joint vision-language INT4 and INT2 quantization. We also study the contributions of offline and on-policy supervision in the target INT2 setting.

\subsection{Experimental Setup}
\label{sec:exp_setup}

\paragraph{Task Suite}
We evaluate the models on four in-house datasets covering practical on-device vision-language applications.

\textbf{Sitting Posture Monitoring (SPM)} classifies the person closest to the camera into correct, incorrect (e.g., slouching or lying on the desk), or non-sitting postures, measured by three-class accuracy.

\textbf{Home Safety Monitoring (HSM)} detects falls, unsafe child behavior, abnormal pet behavior, and fire hazards from household surveillance images.

\textbf{TV--Camera Query Routing (TCR)} comes from a smart-TV assistant with a TV-screen camera and a room-facing camera. Given a multi-turn conversation with previous predictions as history and both images, the model selects no visual input, the TV-screen image, or the room-camera image, measured by per-turn routing accuracy.

\textbf{Mobile GUI Control (MGC)} predicts the next Android operation (e.g., clicking, typing, or swiping) from a user goal, the current screenshot, and previous actions, measured by exact-step accuracy. Coordinate-based actions are accepted when their distance from the target is below 50 in a normalized $1000\times1000$ space.

Table~\ref{tab:datasets} summarizes the training and test data.

\begin{table}[htbp]
\caption{Statistics of the Four On-Device Tasks}
\begin{center}
\setlength{\tabcolsep}{3pt}
\begin{tabular}{lrrl}
\toprule
\textbf{Task} & \textbf{Train} & \textbf{Test} &
\textbf{Primary Metric} \\
\midrule
SPM & 7,353 & 294 & 3-class accuracy \\
HSM & 15,723 & 751 & Label-set accuracy \\
TCR & 11,025 & 500 & Turn accuracy \\
MGC & 58,415 & 2,062 & Exact-step accuracy \\
\bottomrule
\multicolumn{4}{l}{
Train denotes effective examples in one interleaved epoch.
}
\end{tabular}
\label{tab:datasets}
\end{center}
\end{table}

A development subset containing 100 records from each training source is used for quantization calibration and pipeline validation. This subset is excluded from the independent task-specific test sets in Table~\ref{tab:datasets}.

\paragraph{Two-Stage Pipeline}
All experiments use Qwen3-VL-4B~\cite{qwen3vl} as the shared vision-language backbone. The original system contains four task-specific LoRA experts attached to the same BF16 backbone. During evaluation, each input is processed using its corresponding task adapter. We refer to this setting as Multi-LoRA Experts and use task-specific adapter.

In the first stage, the task-specific knowledge is distilled into a single adapter-free model, denoted as Unified-BF16. Concretely, each of the four LoRA experts is attached in turn to the shared BF16 backbone to label its own task data, and one adapter-free model is then trained on the union of these expert-generated responses with the standard next-token cross-entropy objective $\mathcal{L}_{\mathrm{cons}}=-\frac{1}{N}\sum_{t=1}^{N}\log p_{\mathrm{uni}}(y_t^\ast\mid h_t^\ast)$, where $p_{\mathrm{uni}}$ is the consolidated model. Training uses the same four-task data mixture and the same optimizer configuration as the recovery stage (AdamW, learning rate $8\times10^{-6}$, cosine schedule, 100 warm-up steps, effective batch size 8) for three epochs on one A100 GPU. The role of this stage is solely to remove adapter multiplicity: it collapses four checkpoints and their runtime selection logic into one BF16 model at unchanged precision and unchanged backbone memory. The unified model is subsequently used as the teacher and initialization source for low-bit compression. The second stage then addresses the complementary cost that consolidation cannot remove, namely the memory of the shared backbone itself, by converting it into a mixed-precision asymmetric INT2 student while preserving the consolidated multi-task behavior.

In the second stage, Unified-BF16 is directly quantized to obtain the GPTQ-INT4 and GPTQ-INT2 baselines. Starting from the corresponding GPTQ initialization, UniCompress-INT4 and UniCompress-INT2 apply the proposed recovery procedure at the same precision. INT4 provides a moderate-precision reference, while INT2 is our target deployment setting.

\begin{table*}[htbp]
\caption{Results Across the Two-Stage Consolidation and Quantization Pipeline}
\begin{center}
\setlength{\tabcolsep}{7pt}
\begin{tabular}{ccccccc}
\toprule
\textbf{Model} &
\textbf{Size} &
\textbf{SPM} &
\textbf{HSM} &
\textbf{TCR} &
\textbf{MGC} &
\textbf{Avg.} \\
\midrule
Multi-LoRA Experts$^{\mathrm{a}}$
& $\sim$8.9 GB $+$ 0.2 GB adapters
& 94.9\% & 94.5\% & \textbf{97.6\%} & 87.5\% &
93.6\% \\
Unified-BF16$^{\mathrm{b}}$
& $\sim$8.9 GB
& 95.2\% & 94.7\% & 97.2\% & 87.7\% &
\textbf{93.7\%} \\
GPTQ-INT4
& $\sim$3.3 GB
& 93.2\% & 93.7\% & 96.8\% & 87.5\% & 92.8\% \\
UniCompress-INT4
& $\sim$3.3 GB
& \textbf{95.2\%} & 94.5\% & 96.6\% &
\textbf{87.8\%} & 93.5\% \\
GPTQ-INT2
& $\sim$2.1 GB
& 71.8\% & 85.9\% & 58.6\% & 17.8\% & 58.5\% \\
UniCompress-INT2
& $\sim$2.1 GB
& 93.2\% & \textbf{95.1\%} & 97.4\% & 85.4\% &
92.8\% \\
\bottomrule
\end{tabular}

\vspace{2pt}
\begin{minipage}{0.82\textwidth}
$^{\mathrm{a}}$Uses task-specific adapters; the four adapters add about 0.2 GB (roughly 50 MB each in BF16) to the 8.9 GB backbone.
$^{\mathrm{b}}$Consolidated adapter-free model.
GPTQ models are quantized from Unified-BF16, and UniCompress starts from the corresponding GPTQ initialization.
\end{minipage}

\label{tab:main_results}
\end{center}
\end{table*}

\begin{table*}[htbp]
\caption{Ablation Study of the INT2 Training Objectives}
\begin{center}
\setlength{\tabcolsep}{10pt}
\begin{tabular}{lccccc}
\toprule
\textbf{Setting} &
\textbf{SPM} &
\textbf{HSM} &
\textbf{TCR} &
\textbf{MGC} &
\textbf{Avg.} \\
\midrule
Offline only
& 93.2\% & 92.5\% & 95.8\% & 83.9\% & 91.4\% \\
OPD only
& 92.9\% & 93.7\% & 96.4\% & 84.7\% & 91.9\% \\
UniCompress-INT2
& \textbf{93.2\%} & \textbf{95.1\%} & \textbf{97.4\%} &
\textbf{85.4\%} & \textbf{92.8\%} \\
\bottomrule
\end{tabular}
\label{tab:ablation}
\end{center}
\end{table*}

\paragraph{Implementation Details}
For the low-bit models, the projection weights in both the vision encoder and language model are quantized. INT4 uses a group size of 32, while INT2 uses group-wise asymmetric quantization with a group size of 64. Embeddings, normalization layers, and the language-model output remain in BF16. During recovery, the trainable master weights are initialized from the corresponding GPTQ model and dynamically fake-quantized.

Quantization recovery is performed for two epochs using AdamW with a learning rate of $8\times10^{-6}$, a cosine learning-rate scheduler, 100 warm-up steps, and an effective batch size of 8. We set $\beta=0.5$ and $\tau=0.1$ in Eq.~\eqref{eq:weight}, and linearly increase $\lambda(s)$ in Eq.~\eqref{eq:total} from $0$ to $1$ over the first training epoch. Training uses BF16 computation on one NVIDIA A100 80GB GPU with a fixed random seed of 42. All models use the same prompts, image preprocessing, and greedy decoding during evaluation.

For Qwen3-VL-4B, the approximate weight storage is 8.9 GB in BF16, 3.3 GB in INT4, and 2.1 GB in mixed-precision INT2. These values exclude tokenizer and configuration files. The Multi-LoRA setting additionally stores the four task-specific adapters, which occupy about 0.2 GB in total (roughly 50 MB per adapter in BF16), bringing its full checkpoint footprint to about 9.1 GB. Consolidation removes this adapter storage entirely and, because a single weight set then serves all tasks, further permits the KV cache of an ongoing session to be reused across tasks---a benefit that per-task adapter switching cannot provide, since changing the active adapter invalidates the cached states.

For each model, the macro-average score is calculated as the unweighted mean of the four primary task metrics.

\subsection{Main Results}
\label{sec:main_results}

\begin{figure}[t]
\centerline{\includegraphics[width=\columnwidth]{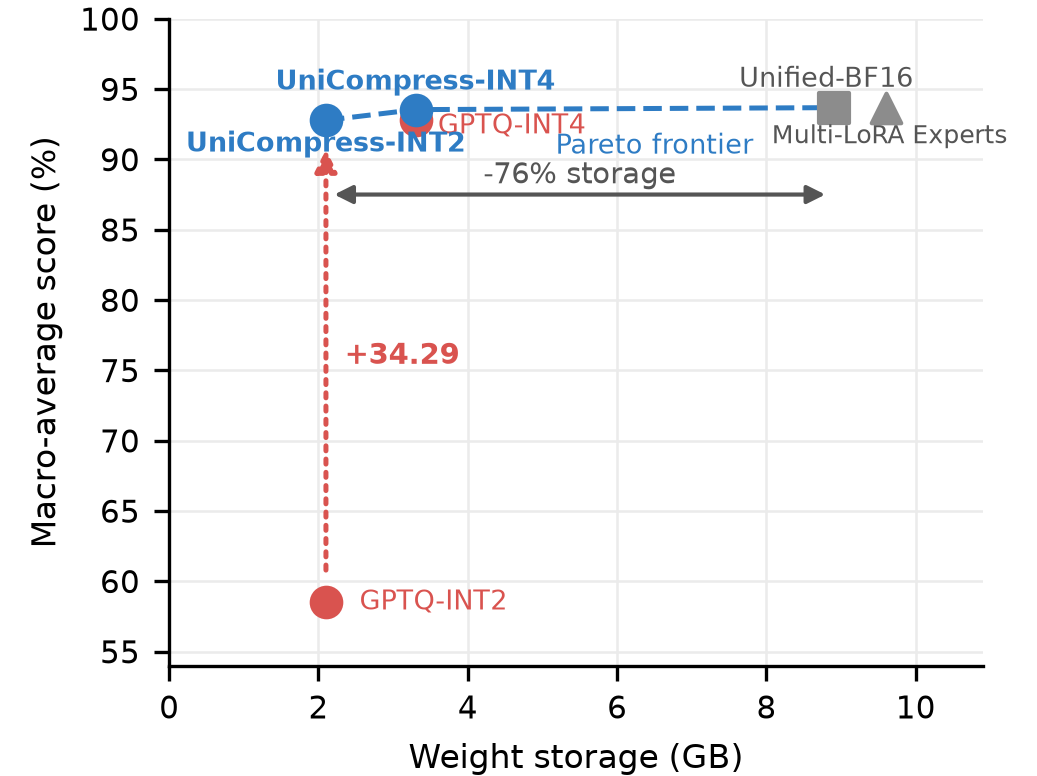}}
\caption{Weight storage versus macro-average score. UniCompress-INT2 lies on the Pareto frontier: it recovers 34.3 points over GPTQ-INT2 at the same 2.1 GB storage and reduces the weight storage by 76\% relative to BF16 models.}
\label{fig:pareto}
\end{figure}

Table~\ref{tab:main_results} reports the results across the complete two-stage pipeline. All task scores are percentages, and model sizes denote approximate weight storage. Fig.~\ref{fig:pareto} summarizes the resulting storage--accuracy trade-off.

\paragraph{Multi-LoRA Consolidation}
The Multi-LoRA Experts achieve a macro-average score of 93.6\%. After consolidating the four task-specific adapters, Unified-BF16 obtains 93.7\%, showing that the first stage introduces no overall performance loss. Task-level scores remain within 0.4 points of the corresponding experts, indicating that the individual expert capabilities are preserved within a single adapter-free model.

\paragraph{INT4 Quantization}
Direct GPTQ-INT4 reduces the macro-average score from 93.7\% to 92.8\%, corresponding to a loss of 0.9 points. UniCompress-INT4 raises the score to 93.5\%, recovering 0.7 points and leaving a gap of only 0.2 points to Unified-BF16. It also achieves the best SPM and MGC results among all evaluated models. These results show that moderate-precision quantization can preserve the consolidated capabilities with relatively limited recovery. Meanwhile, INT4 reduces the approximate weight storage from 8.9 GB to 3.3 GB.

\paragraph{INT2 Quantization}
Direct INT2 quantization is considerably more destructive. GPTQ-INT2 obtains only 58.5\%, a reduction of 35.2 points from Unified-BF16. The largest degradation occurs on MGC, where accuracy drops from 87.7\% to 17.8\%, followed by TCR, which drops from 97.2\% to 58.6\%.

UniCompress-INT2 improves the macro-average score to 92.8\%, recovering 34.3 percentage points over GPTQ-INT2. The final gap is only 0.9 points to Unified-BF16 and 0.8 points to the original Multi-LoRA Experts. The largest gains appear on TCR (+38.8) and MGC (+67.6), supporting the need to recover behavior on the quantized model.

UniCompress-INT2 also slightly exceeds Unified-BF16 on HSM and TCR, while remaining 2.0 and 2.3 points lower on SPM and MGC. Overall, the final INT2 model retains most of the capabilities of both the unified model and the original task-specific experts while reducing the approximate weight storage from 8.9 GB to 2.1 GB.

\subsection{Ablation Study}
\label{sec:ablation}

We study the two training signals used in the INT2 recovery stage. Offline only trains on annotated sequences without student rollout. OPD only uses quantized-student trajectories without the offline anchoring branch. UniCompress-INT2 combines both forms of supervision. Table~\ref{tab:ablation} reports the results.

OPD-only training (91.9\%) outperforms offline-only training (91.4\%) on HSM, TCR, and MGC, suggesting that student-generated trajectories are particularly useful where generation errors accumulate. Combining both objectives achieves the best result on all four tasks (92.8\%), indicating that offline examples provide a stable multi-task knowledge anchor while on-policy trajectories expose the student to states caused by its own INT2 predictions.

\section{Conclusion}
\label{sec:conclusion}

We presented UniCompress, a two-stage framework for converting multiple task-specific LoRA experts into a single compact vision--language model. UniCompress first consolidates the experts into a unified Qwen3-VL-4B model and then jointly quantizes and recovers the projection weights in its vision encoder and language backbone. Across four on-device tasks, Unified-BF16 preserves the capabilities of the original experts, while UniCompress-INT2 improves the macro-average score by 34.3 percentage points over GPTQ-INT2 and remains within 0.9 points of Unified-BF16 at only 2.1 GB of weight storage. These results show that deployment-faithful recovery can preserve heterogeneous task capabilities under aggressive joint vision--language quantization, enabling a single adapter-free INT2 model for resource-constrained edge deployment.

\section*{Acknowledgment}
This work was supported by the AI Technology Center of Juhaokan Technology Co., Ltd.

\end{document}